# An Implementation of a Method
# for Computing the Uncertainty
# in Inferred Probabilities in Belief Networks

**Peter Che**
Computer Science Department
Illinois Institute of Technology
10 West 31st Street
Chicago, IL 60616

**Richard E. Neapolitan**
Computer Science Department
Northeastern Illinois University
5500 North St. Louis Avenue
Chicago, IL 60625

**James Kenevan & Martha Evens**
Computer Science Department
Illinois Institute of Technology
10 West 31st Street
Chicago, IL 60616

## Abstract

In recent years the belief network has been used increasingly to model systems in AI that must perform uncertain inference. The development of efficient algorithms for probabilistic inference in belief networks has been a focus of much research in AI. Efficient algorithms for certain classes of belief networks have been developed, but the problem of reporting the uncertainty in inferred probabilities has received little attention. A system should not only be capable of reporting the values of inferred probabilities and/or the favorable choices of a decision; it should report the range of possible error in the inferred probabilities and/or choices. Two methods have been developed and implemented for determining the variance in inferred probabilities in belief networks. These methods, the Approximate Propagation Method and the Monte Carlo Integration Method are discussed and compared in this paper.

## 1 INTRODUCTION

A belief network consists of a directed acyclic graph, $(V, E)$, where $V$ is a set of vertices and $E$ is a set of edges, in which each $v$ in $V$ represents a set of mutually exclusive and exhaustive alternatives, along with a joint probability distribution P on the alternatives of the nodes in $V$. Each node in $V$ is called a chance node because it represents the possible outcomes of a chance occurrence. The fundamental assumption in a belief network is that the value assumed by a node is probabilistically independent of the values assumed by all other nodes in the network, except the descendents of the given values of the parents of the node. It can be shown, given this restriction on $P$, that $P$ can be retrieved from the product of the conditional distributions of each node given the values of its parents. Thus it is only necessary to specify these conditional distributions. Discussions concerning how inference can be used as the underlying structure in expert systems that perform uncertain inference can be found in [Neapolitan 1990] and [Pearl 1988].

The problem of performing both probability propagation and abductive inference in an arbitrary belief network has been proven to be NP-hard [Cooper 1990]. That is to say, there is no single algorithm that is efficient for all belief networks. Many efficient algorithms have been developed for special kinds of belief networks and applications, including probability propagation in singly connected networks by Pearl [1988] and probability propagation in a tree of cliques by Lauritzen and Spiegelhalter [1988]. Efficient approximation methods [Chavez & Cooper 1990a, 1990b] have also been developed. However, these methods have been developed based on the scenario of point probabilities with precise values. Lopez [1990] claims that the major shortcoming of the probabilistic approach is the assumption that all the probabilities are specified precisely, but in practice they reflect subjective judgments that are inherently imprecise. Spiegelhalter [1989] also points out that the imprecision of a single valued "point" probability can be caused by imprecise assessments and that the probability value is very sensitive to further relevant information.

It is important to determine the uncertainty in inferred probabilities in belief networks. A system should not only report the values of point probabilities, but should also report the uncertainty in the probabilities [Neapolitan 1993]. Knowledge of the uncertainty in inferred probabilities makes the decision maker aware of the quality of the probabilities and helps him decide whether additional information should be acquired.

## 2 STATISTICAL VARIANCE AND DIRICHLET DISTRIBUTIONS

The variance of a probability value is clearly a good candidate for representing the uncertainty in the point probability. When the probability distributions are Dirichlet, the values $E(P_i)$, $E(P_i^2)$, and $E(P_iP_j)$ can be derived as follows:

$$E(P_i) = \frac{a_i + 1}{\sum_{k=1}^{t} a_k + t}$$

$$E(P_i^2) = \frac{a_i + 2}{\sum_{k=1}^{t} a_k + t + 1} E(P_i)$$

$$E(P_iP_j) = \frac{(a_i + 1)(a_j + 1)}{(\sum_{k=1}^{t} a_k + t + 1)(\sum_{k=1}^{t} a_k + t)}$$

The variance of a probability $P_i$ is represented as follows.

$$V(P_i) = E(P_i^2) - (E(P_i))^2$$

$E(P_i)$, $E(P_i^2)$, and $E(P_iP_j)$ are the three main factors for calculating the variance of the probabilities in a belief network in the method that is developed in [Neapolitan 1991]. ($P_i$ and $P_j$ are random variables for the ith and jth alternatives of the same node.)

## 3 THE APPROXIMATE PROPAGATION METHOD

Assume in a singly connected network, $AF_1$, ..., $E$, ..., $AF_n$ are parents of $F$, $CF_1$, ..., $G$, ..., $CF_n$ are children of $F$, $AG_1$, ..., $F$, ..., $AG_n$ are parents of $G$, and $CG_1$, ..., $H$, ..., $CG_n$ are children of $G$. The initial marginal value of $E(P_i)$, $E(P_i^2)$, and $E(P_iP_j)$ for each non-root node in singly connected networks can be derived as follows:

$$E[P(f^t)]$$
$$= \alpha \sum_{t_{af_1}, \dots, t_{af_n}} \cdot P(f^t \mid af_1{}^{t_{af_1}}, \dots, af_n{}^{t_{af_n}})$$
$$\prod_i^n P(af_i{}^{t_{af_i}})$$
$$E[P(f^s)P(f^t)]$$
$$= \alpha^2 \sum_{s_{af_1}, \dots, s_{af_n}, t_{af_1}, \dots, t_{af_n}}$$
$$E[P(f^s \mid af_1{}^{s_{af_1}}, \dots, af_n{}^{s_{af_n}})$$
$$P(f^t \mid af_1{}^{t_{af_1}}, \dots, af_n{}^{t_{af_n}})]$$
$$\prod_{i,j}^n E[P(af_i{}^{s_{af_i}})P(af_j{}^{t_{af_j}})]$$

(where $af_i{}^{t_{af_i}}$ is an alternative of node $AF_i$).
The terms

$$E[P(af_i{}^{s_{af_i}})P(af_j{}^{s_{af_j}})]$$

and

$$E[P(f^s \mid af_1{}^{s_{af_1}}, \dots, af_n{}^{s_{af_n}})$$

$$P(f^t \mid af_1{}^{t_{af_1}}, \dots, af_n{}^{t_{af_n}})]$$

are stored in the network as $E(P_iP_j)$.

When new evidence is observed, the distribution of probability and variance in the belief network must be updated. This update procedure can be achieved by local computation and passing messages among neighboring nodes in the belief network [Neapolitan 1991; Che 1992]. A generalized algorithm for the propagation of variance in the singly connected network is as follows. Assume:

$$
\begin{aligned}
\pi(f^t) &= \pi \text{ message to } F \text{ (alternative } t) \\
&\quad \text{ from all parents of } F \\
\pi_{-F}(e^t) &= \pi \text{ message from parent } E \\
&\quad \text{ (alternative } t) \text{ to child } F \\
\lambda(f^t) &= \lambda \text{ message to } F \text{ (alternative } t) \\
&\quad \text{ from all children of } F \\
\lambda_{-G}(f^t) &= \lambda \text{ message from child } G \\
&\quad \text{ to parent } F \text{ (alternative } t) \\
\pi'(f^s, f^t) &= \pi' \text{ message to } F \text{ (alt } s, t) \\
&\quad \text{ from all parents of } F \\
\pi'_{-F}(e^s, e^t) &= \pi' \text{ message from parent } E \\
&\quad \text{ (alt } s, t) \text{ to child } F \\
\lambda'(f^s, f^t) &= \lambda' \text{ message to } F \text{ (alt } s, t) \\
&\quad \text{ from all children of } F \\
\lambda'_{-G}(f^s, f^t) &= \lambda' \text{ message from child } G \\
&\quad \text{ to parent } F \text{ (alt } s, t)
\end{aligned}
$$

The variance can be derived as follows:

$$V[P(f^t)] = E[P(f^t)^2] - (E[P(f^t)])^2$$

$$E[P(f^t)] = \alpha \lambda(f^t) \pi(f^t)$$

$$\pi(f^t) = \sum_{t_{af_1}, \dots, t_{af_n}} P(f^t \mid af_1{}^{t_{af_1}}, \dots, af_n{}^{t_{af_n}})$$

$$\prod_i \pi_{-F}(af_i{}^{t_{af_i}})$$

$$\pi_{-F}(e^t) = \pi(e^t)\prod_{k:CE_k \neq F}\lambda_{-CE_k}(e^t)$$

$$\lambda(f^t) = \prod_j \lambda_{-CF_j}(f^t)$$

$$\lambda_{-G}(f^t) = \sum_u [\sum_{u_{ag_1},\dots,u_{ag_n}:AG_i \neq F}$$

$$P(g^u \mid ag_1^{u_{ag_1}},\dots,f^t,\dots,ag_n^{u_{ag_n}})$$

$$\prod_{k:AG_k \neq F}\pi_{-G}(ag_k^{u_{ag_k}}\prod_j \lambda_{-CG_j}(g^u)]$$

$$E[P(f^s)P(f^t)] = \alpha^2 E[\lambda'(f^s,f^t)\pi'(f^s,f^t)]$$

$$\begin{aligned}
\pi'(f^s,f^t) &= \sum_{s_{af_1},\dots,s_{af_n},t_{af_1},\dots,t_{af_n}}\\
&\quad E[P(f^s \mid af_1^{s_{af_1}},\dots,af_n^{s_{af_n}})\\
&\quad P(f^t \mid af_1^{t_{af_1}},\dots,af_n^{t_{af_n}})]\\
&\quad \prod_i \pi'_{-F}(af_i^{s_{af_i}},af_i^{t_{af_i}})
\end{aligned}$$

$$\pi'_{-F}(e^s,e^t) = \pi'(e^s,e^t)\prod_{k:CE_k \neq F}\lambda'_{-CE_k}(e^s,e^t)$$

$$\lambda'(f^s,f^t) = \prod_j \lambda'_{-CF_j}(f^s,f^t)$$

$$\lambda'_{-G}(f^s,f^t) = \sum_{u,v}[\sum_{u_{ag_1},\dots,u_{ag_n},v_{ag_1},\dots,v_{ag_n}}$$

$$E[P(g^u \mid ag_1^{u_{ag_1}},\dots,f^s,\dots,ag_n^{u_{ag_n}})$$

$$P(g^v \mid ag_1^{v_{ag_1}},\dots,f^t,\dots,ag_n^{v_{ag_n}})]$$

$$\prod_{k:AG_k \neq F}\pi'_{-G}(ag_k^{u_{ag_k}},ag_k^{v_{ag_k}})$$

$$\prod_j \lambda'_{-CG_j}(g^u,g^v)]$$

(where $CE_i$ is a child of $E$, $CG_i$ is a child of $G$, $AG_i$ is a parent of $G$, and $ag_i^{u_{ag_i}}$ is an alternative of $AG_i$).

The terms

$$E[P(f^s \mid af_1^{s_{af_1}},\dots,af_n^{s_{af_n}})$$

$$P(f^t \mid af_1^{t_{af_1}},\dots,af_n^{t_{af_n}})]$$

and

$$E[P(g^u \mid ag_1^{u_{ag_1}},\dots,f^s,\dots,ag_n^{u_{ag_n}})$$

$$P(g^v \mid ag_1^{v_{ag_1}},\dots,f^t,\dots,ag_n^{v_{ag_n}})]$$

are stored in the network as $E(P_iP_j)$.

As in the version of Pearl's [1986] probability propagation method described in [Neapolitan 1990], initially all lambda values are set to 1 and pi values are calculated from top down throughout the network. When a variable is instantiated, a new set of lambda and pi messages are sent to all its parent and child nodes. The messages are then propagated through the entire network until a new balance of probability and variance distributions are reached.

## 4 THE MONTE CARLO INTEGRATION METHOD

In the Monte Carlo Integration Method [Neapolitan & Kenevan 1990; Che 1992] random samples of the probability distribution in a belief network are generated, the update of the probability distribution is computed for each sample, and the variance is derived when a sufficiently large sample is collected. A very long processing time is necessary if the demand for accuracy is high and the size of the network is large. The numerical integration [Kincaid 1985] of the expected value $E(P(f^i \mid W)^2)$ can be derived as follows:

$$\begin{aligned}
&E(P(f^i \mid W)^2)\\
&= \int_U P(f^i \mid W,U)^2 dP(U \mid W)\\
&= \frac{1}{P(W)}\int_U P(f^i \mid W,U)^2 P(W \mid U)dP(U)\\
&= \frac{1}{P(W)}(\frac{1}{t}\sum_{j=1}^t P(f^i \mid W,U_j)^2 P(W \mid U_j))
\end{aligned}$$

where $W$ is a set of instantiated nodes and $F$ is the node of interest in the belief network. The value $t$ is a constant; a larger $t$ value implies a smaller error in the result of numeric integration. The probabilities $P(f^i \mid W,U_j)$ and $P(W \mid U_j)$ can be derived by random sampling as follows:

$$r = \int_0^p u(x)dx$$

where $r \in (0,1)$ is a random number and $u(x)$ is a density function.

## 5 COMPARISON OF METHODS

The results in the following examples show that the posterior variances derived by using the Approximate Propagation Method and the Monte Carlo Integration method become very close when there is a reasonable amount of certainty in prior probabilities. In the examples in Tables 1, 2, and 3 we assume that all propositional variables have two alternatives and all prior and conditional probabilities are equal to 0.5. The letter $a$ represents the specified value in the Dirichlet distribution of the value of each point probability. In Table 1 we assume that the propositional variable $E$ is a single parent of $F$. In Table 2 we assume that propositional variable $E$ is a single parent of $F$, and $F$ is a single parent of $G$. In Table 3 we assume that the propositional variable $E$ is a single parent of $F$ and $G$.

Table 1: The Expected Values $E(P(e_i \mid f_j)^2)$
When a Single Child Node is Instantiated
(The second and third columns contain values
calculated by the Monte Carlo Integration and the
Approximate Propagation Methods)

| a | MCIM | APM | Prior |
|---|---|---|---|
| 0 | 0.360 | 0.444 | 0.333 |
| 1 | 0.319 | 0.360 | 0.300 |
| 2 | 0.300 | 0.327 | 0.286 |
| 5 | 0.278 | 0.290 | 0.269 |
| 10 | 0.266 | 0.272 | 0.261 |
| 20 | 0.260 | 0.262 | 0.256 |

The results in Figures 1 and 2 show that when the number of instantiated child nodes increases, the variance in the parent node increases quickly. The increase of variance is faster in the Approximation Method than in the Monte Carlo Integration Method, especially when the certainty in prior probabilities is low. However, when there is reasonable certainty in the prior probabilities and the number of instantiated child nodes is not very large, the resulting variances from the two methods are very close.

Table 2: The Expected Values $E(P(e_i \mid f_j)^2)$
When Two Child Nodes Are Instantiated
(The second and third columns contain values
calculated by the Monte Carlo Integration and the
Approximate Propagation Methods)

| a | MCIM | APM |
|---|---|---|
| 0 | 0.374 | 0.593 |
| 1 | 0.329 | 0.432 |
| 2 | 0.310 | 0.373 |
| 5 | 0.282 | 0.312 |
| 10 | 0.268 | 0.285 |
| 20 | 0.260 | 0.268 |

Table 3: The Expected Values $E(P(e_i \mid f_j)^2)$
When a Single Grandchild Node Is Instantiated

| a | MCIM | APM |
|---|---|---|
| 0 | 0.324 | 0.407 |
| 1 | 0.298 | 0.336 |
| 2 | 0.280 | 0.309 |
| 5 | 0.265 | 0.280 |
| 10 | 0.260 | 0.267 |
| 20 | 0.255 | 0.259 |

The results in Figures 1 and 2 show that when the number of instantiated child nodes increases, the variance in the parent node increases quickly. The increase of variance is faster in the Approximation Method than in the Monte Carlo Integration method, especially when the certainty in prior probabilities is low. However, when there is reasonable certainty in the prior probabilities and the number of instantiated child nodes is not very large, the resulting variances from the two methods are very close.

In the examples in Figures 1 and 2 we assume that the propositional variable $E$ is the root node, and $E$ has child nodes $C_1, \ldots, C_j, \ldots, C_n$. All propositional variables have two alternatives. The letters $a$ and $b$ represent the specified values in the Dirichlet distributions of the probabilities in the belief networks.

The results in Figures 3 and 4 show that when the level of instantiated descendant nodes becomes deeper, the variance in the root node reaches a constant. The value of the variance is greater in the Approximate Propagation Method than that in the Monte Carlo Integration Method. When there is reasonable certainty in the prior probabilities, the variances from the two methods become very close.

In the examples in Figures 3 and 4 we assume that the propositional variable $E$ is the root node and $L$ is the leaf node in a chain. All propositional variables have two alternatives. The letters $a$ and $b$ represent the specified values in the Dirichlet distributions of the probabilities in the belief networks.

# 6 DISCUSSION AND CONCLUSION

The results in the above examples show that the posterior variances derived by using the Approximate Propagation Method is always larger than obtained from the Monte Carlo Integration Method. When the certainty in the prior probability decreases the difference in posterior variances derived by using the two methods become larger. When the certainty in the prior probability increases the posterior variances derived by using the two methods become closer. When the certainty in the prior probabilities is above a certain level (for example, $a \geq 10$ and $b \geq 10$) the posterior variances derived by using both methods become very close.

When the network becomes large, it can take a very long time to obtain accurate values of the variances using the Monte Carlo Integration Method. This situation is similar to the slow convergence problem in the Stochastic Simulation method. Tradeoffs must be made between the accuracy of the result and the length of time to generate the result. On the other hand, the Approximate Propagation Method is very efficient, especially for large networks, in comparison with the Monte Carlo Integration Method. The average running time on a 386/SX PC for networks with less than twenty nodes and two alternatives for each propostional variable is a few seconds.

## References

Chavez, M.R., and Cooper, G.F. (1990a). An Empirical Evaluation of a Randomized Algorithm for Probabilistic Inference. *Uncertainty in Artificial Intelligence 5,* M. Henrion, R.D. Shachter, L.N. Kanal and J.F. Lemmer, eds., North Holland, Amsterdam: 191-206.

Chavez, M.R., and Cooper, G.F. (1990b). A Randomized Approximation Algorithm for Probabilistic Inference on Bayesian Belief Networks. *Networks,* Vol. 20: 661-685.

Che, P. (1992). *Propagation of Variance of Probabilities in Belief Networks for Expert Systems and Decision Analysis Applications,* Unpublished Ph.D. thesis, Department of Computer Science, Illinois Institute of Technology, Chicago, Illinois.

Cooper, G.F. (1990). The Computational Complexity of Probabilistic Inference Using Bayesian Belief Networks. *Artificial Intelligence,* Vol. 33: 393-405.

Kincaid, C. (1985). *Numerical Mathematics and Computing,* Brooks/Cole Publishing Company, Monterey, California.

Lauritzen, S.L., and Spiegelhalter, D.J. (1988). Local Computations with Probabilities on Graphical Structures and Their Applications to Expert Systems, *Journal of the Royal Statistical Society B,* Vol. 50, No. 2: 157-224.

Lopez de Mantaras, R. (1990). *Approximate Reasoning Models,* Halsted Press, New York, New York.

Neapolitan, R. E. (1990). *Probabilistic Reasoning in Expert Systems, Theory and Algorithms,* Wiley, New York, New York.

Neapolitan, R.E. (1991). Propagation of Variance in Belief Networks, *Proceedings of SPIE Conference on Application of Artificial Intelligence,* Orlando, Florida, April, 1991.

Neapolitan, R.E. (1993). Computing the Confidence in a Medical Decision Obtained from an Influence Diagram. To appear in *Artificial Intelligence in Medicine.*

Neapolitan, R.E., and Kenevan, J.R. (1990). Computation of Variance in Causal Networks. *Proceedings of the Sixth Conference on Uncertainty in Artificial Intelligence,* MIT, Cambridge, Massachusetts: 194-203.

Pearl, J. (1986), Fusion, Propagation, and Structuring in Belief Networks. *Artificial Intelligence,* Vol. 29: 241-288.

Pearl, J. (1988), *Probabilitistic Reasoning in Intelligent Systems,* Morgan Kaufmann, San Mateo, California.

Spiegelhalter, D.J. (1989), A Unified Approach to Imprecision and Sensitivity of Beliefs in Expert Systems. *Uncertainty in Artificial Intelligence 3,* L. N. Kanal and J. F. Lemmer, eds., North Holland, Amsterdam: 199-209.

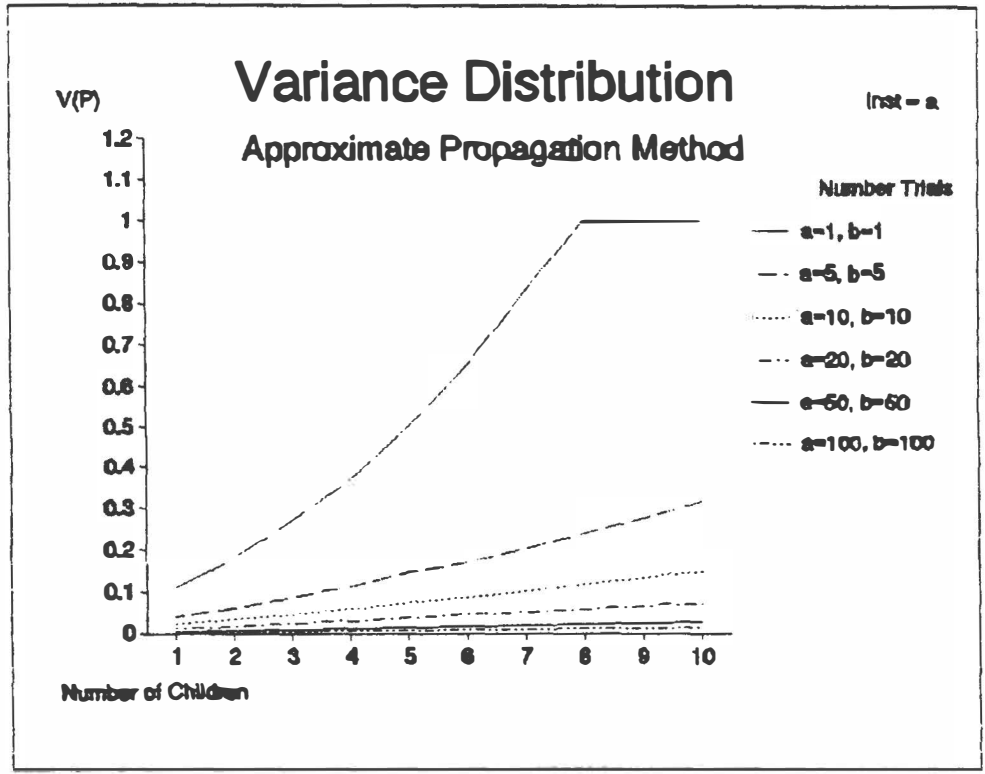

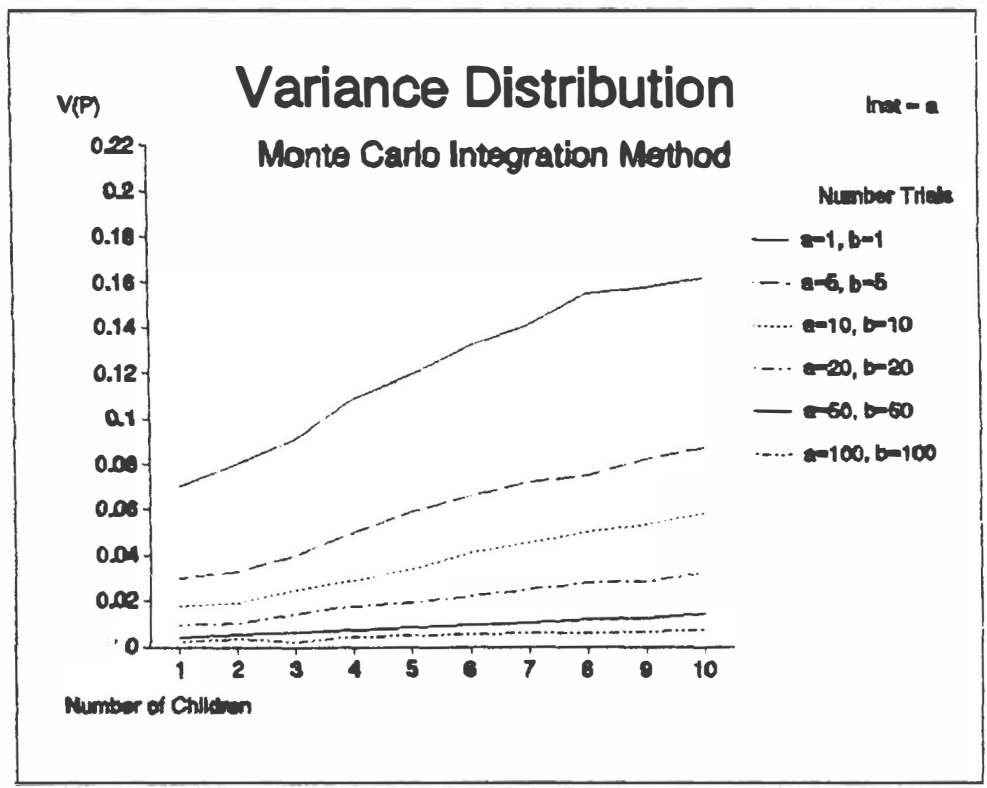

Figure 1: Distribution of Variance by Instantiation of Direct Child Nodes
(assuming that all prior and conditional probabilities are 0.5).

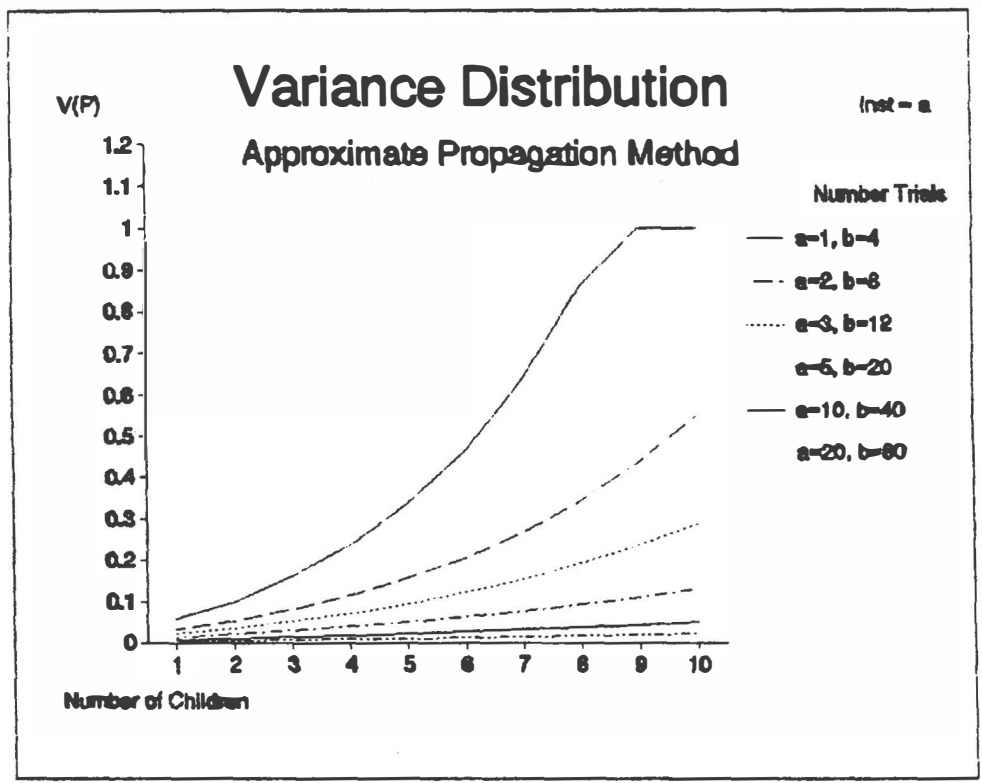

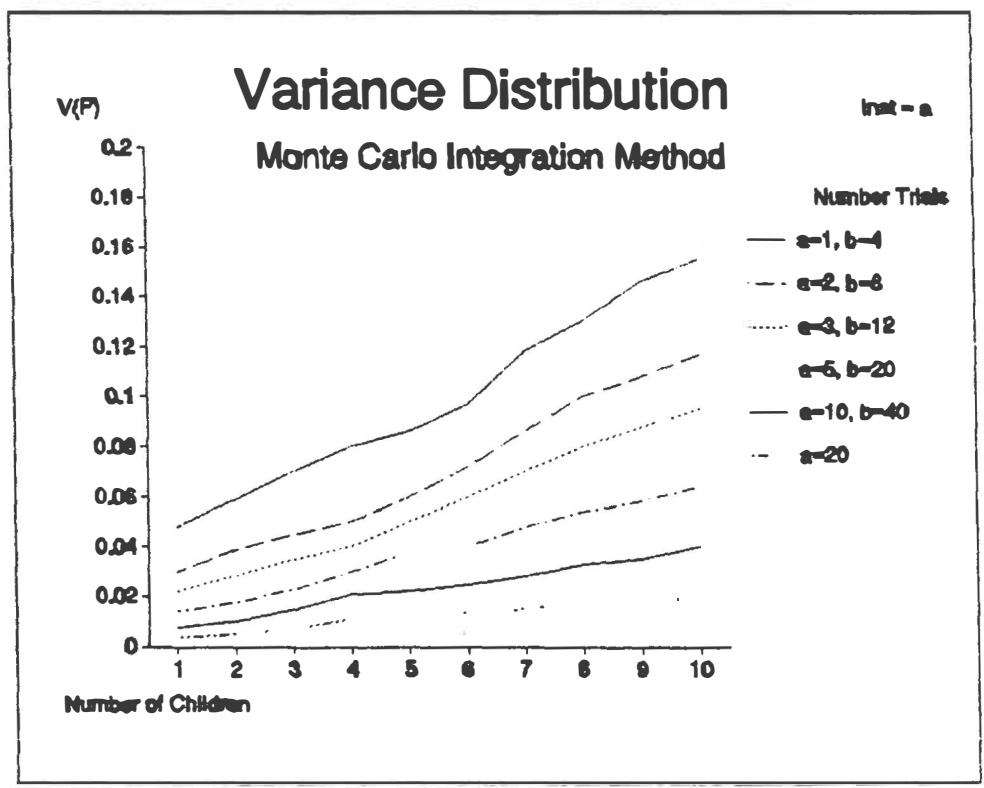

Figure 2: Distribution of Variance by Instantiation of Direct Child Nodes
(assuming that all prior and conditional probabilities are equal to 0.2 and 0.8).

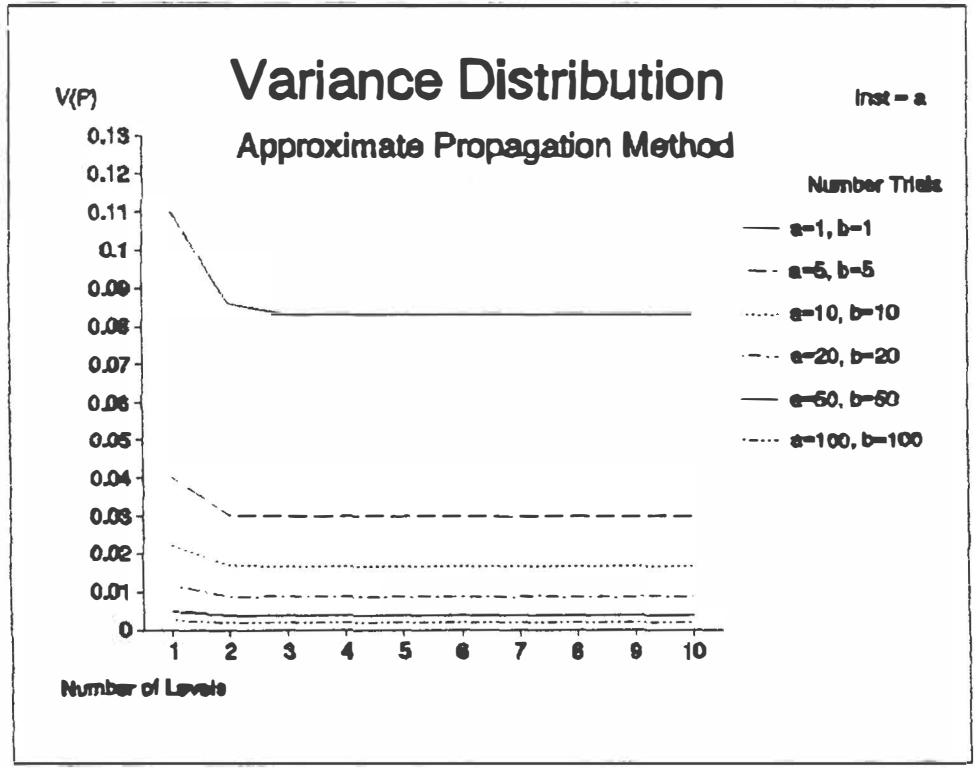

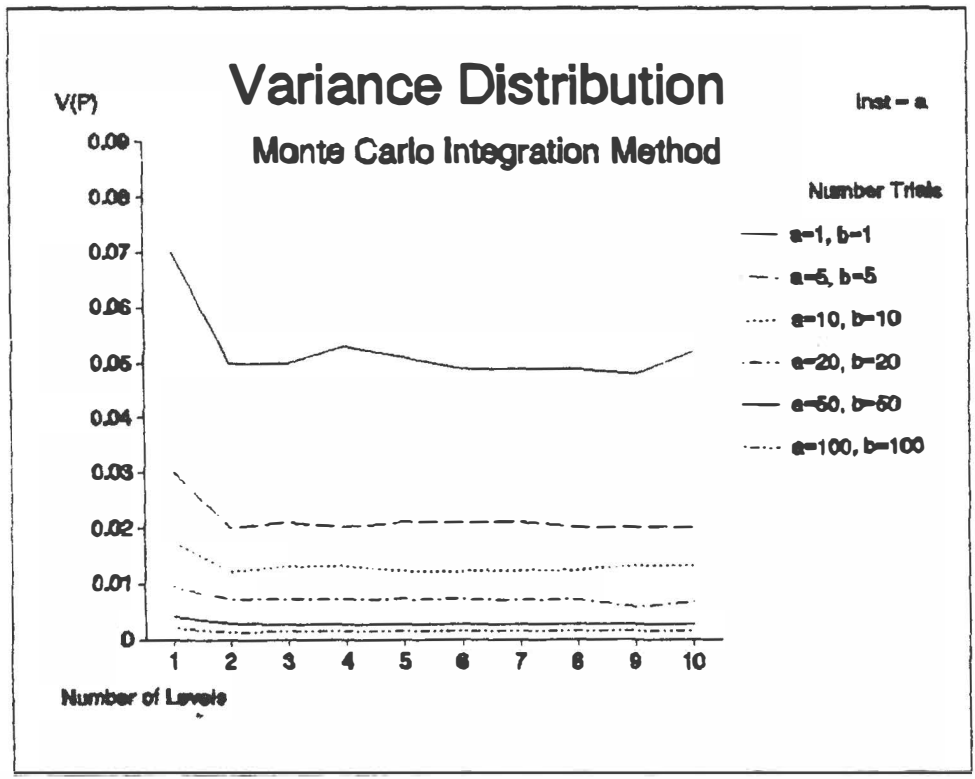

Figure 3: Distribution of Variance by Instantiation of a Leaf Node
(assuming that all prior and conditional probabilities are equal to 0.5).

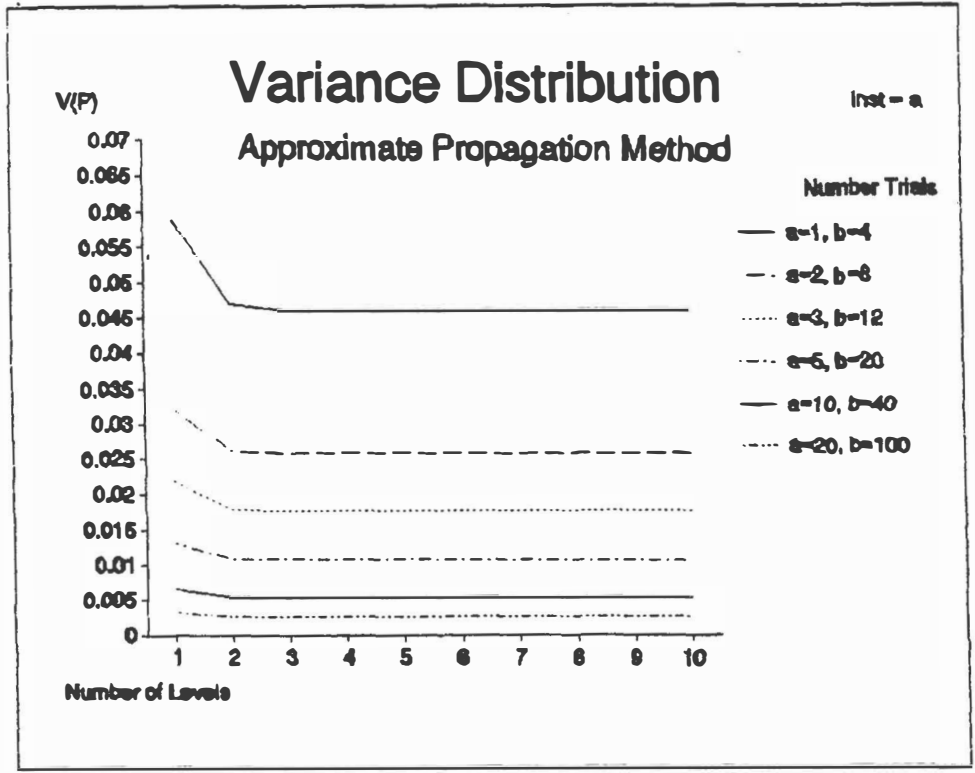

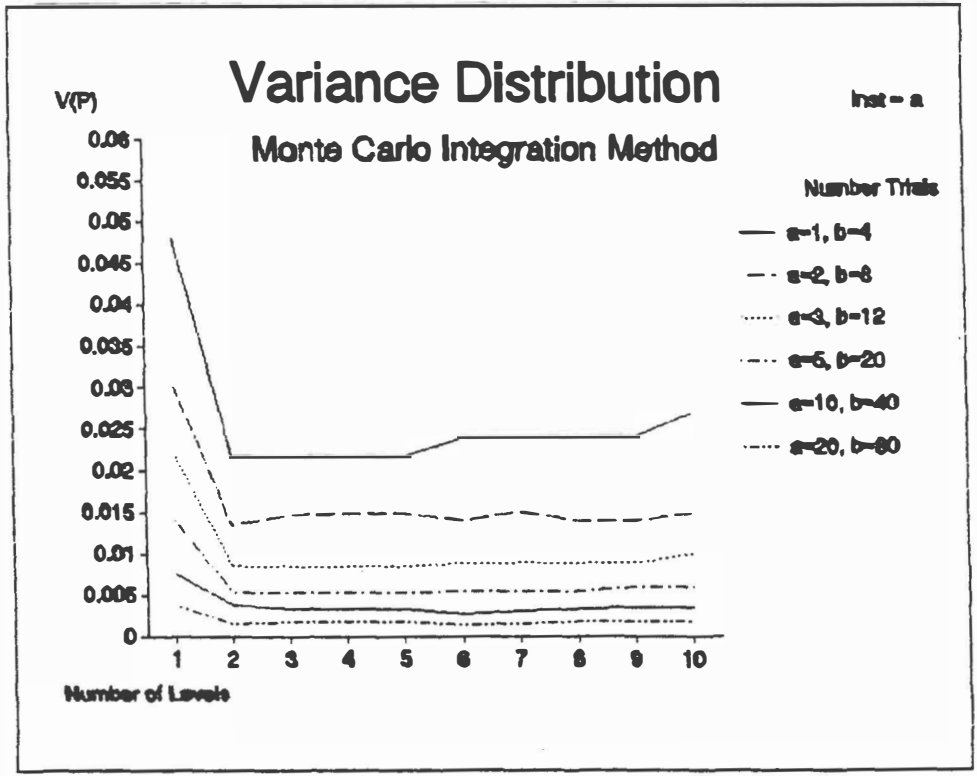

Figure 4: Distribution of Variance by Instantiation of a Leaf Node
(assuming that all prior and conditional probabilities are 0.2 and 0.8).